\documentclass[conference]{IEEEtran}
\IEEEoverridecommandlockouts
\usepackage{cite}
\usepackage{amsmath,amssymb,amsfonts}
\usepackage{algorithmic}
\usepackage{graphicx}
\usepackage{textcomp}
\usepackage{xcolor}

\usepackage{tabularx}
\usepackage{multirow}
\usepackage{booktabs} 
\usepackage{bbding}
\usepackage{array}

\def\BibTeX{{\rm B\kern-.05em{\sc i\kern-.025em b}\kern-.08em
    T\kern-.1667em\lower.7ex\hbox{E}\kern-.125emX}}
\begin{document}

\title{BAGNet: A Boundary-Aware Graph Attention Network for 3D Point Cloud Semantic Segmentation\\
}

\author{\IEEEauthorblockN{
    Wei Tao$^{\spadesuit}$$^{\heartsuit}$,
    Xiaoyang Qu$^{\spadesuit*}$,
    Kai Lu$^{\heartsuit*}$\thanks{$^{*}$Xiaoyang Qu (email: quxiaoy@gmail.com) and Kai Lu (email: kailu@hust.edu.cn) are the corresponding authors.},
    Jiguang Wan$^{\heartsuit}$,
    Shenglin He$^{\heartsuit}$,
    Jianzong Wang$^{\spadesuit}$}
    \IEEEauthorblockA{$^{\spadesuit}$Ping An Technology (Shenzhen) Co., Ltd., Shenzhen, China}
    \IEEEauthorblockA{$^{\heartsuit}$Huazhong University of Science and Technology, Wuhan, China}
}

\maketitle

\begin{abstract}
Since the point cloud data is inherently irregular and unstructured, point cloud semantic segmentation has always been a challenging task. The graph-based method attempts to model the irregular point cloud by representing it as a graph; however, this approach incurs substantial computational cost due to the necessity of constructing a graph for every point within a large-scale point cloud. In this paper, we observe that boundary points possess more intricate spatial structural information and develop a novel graph attention network known as the Boundary-Aware Graph attention Network (BAGNet). On one hand, BAGNet contains a boundary-aware graph attention layer (BAGLayer), which employs edge vertex fusion and attention coefficients to capture features of boundary points, reducing the computation time. On the other hand, BAGNet employs a lightweight attention pooling layer to extract the global feature of the point cloud to maintain model accuracy. Extensive experiments on standard datasets demonstrate that BAGNet outperforms state-of-the-art methods in point cloud semantic segmentation with higher accuracy and less inference time.

\end{abstract}


\section{Introduction}
Point cloud data is prevalent in many application fields, such as 3D shape modeling, action recognition, autonomous driving, and semantic segmentation\cite{watanabe2022point, tao2025PointActionCLIP,he2024prenet, li20203d}. Therefore, many researchers have begun to study point cloud signal processing and analysis. Point cloud semantic segmentation aims to output a category for each point in the point cloud, which plays an essential role in 3D understanding.

Due to the point cloud data's irregular and unstructured characteristics, it is hard to extract semantic features in point cloud semantic segmentation tasks. Some researchers try to convert point cloud data into regular representations to take advantage of traditional convolution neural network (CNN) \cite{qi2017pointnet, qi2017pointnet++}. However, due to the isotropy of the convolution kernel, such methods do not perform well in boundary point semantic segmentation. In order to solve this problem, the graph-based method represents the point cloud as a graph and then uses graph neural network (GNN) to extract features\cite{chen2021gapointnet,wang2019dynamic,lei2020spherical,zhou2021adaptive,zhang2023improving}. These methods enhance the correlation and difference between points and pay different attention to each neighbor point. Although these graph-based methods can improve considerable accuracy, they will cost a lot of computation due to constructing a graph for every point in those large-scale point cloud data.

\begin{figure}
    \centering
    \includegraphics[width=0.48\textwidth]{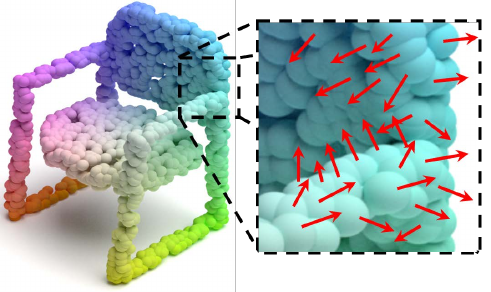}
    \caption{The distribution of normal vectors of different points, where the normal vector distribution of points located at the semantic segmentation boundary is mostly chaotic, and the normal vector distribution of non-boundary points is parallel to their neighbors.}
    \label{fig:challenge}
\end{figure}

We depict the normal vectors of different points in Figure \ref{fig:challenge}. From the figure, we can see that the distribution of normal vectors for points around the semantic segmentation boundary is highly irregular, but the normal vectors of non-boundary points are nearly parallel. This implies that boundary points possess more intricate spatial structural information, making their accurate semantic segmentation challenging, but non-boundary points exhibit simple structures and are thus easier to segment. Leveraging this spatial characteristic of the point cloud, we can concentrate computational resources on the boundary points. By utilizing sophisticated neural networks to extract features from these boundary points, we can enhance semantic segmentation speed and accuracy concurrently.

Based on the above observations, we propose a novel graph attention network called Boundary-Aware Graph attention Network (BAGNet) to address point cloud semantic segmentation tasks. On one hand, BAGNet designs a boundary-aware graph attention layer (BAGLayer). The BAGLayer represents boundary points and their neighbors as a graph, utilizing the fusion of edge-vertex information within the graph to capture contextual attention features. It also enables each boundary point to calculate a unique set of attention coefficients, which encapsulates the connectivity and relational data between the boundary point and all its adjacent neighbors. For other non-boundary points, BAGNet simply uses pointwise multilayer layer perceptron (MLP) \cite{qi2017pointnet++} to extract their features. By using the BAGLayer, BAGNet concentrates most of the computational resources solely on the intricate boundary points, significantly reducing the computation time.
On the other hand, since accurate point semantic segmentation requires not only local features of boundary points but also an appropriate global feature, BAGNet also proposes a lightweight attention pooling layer to extract global features from the raw point cloud to maintain model accuracy.

Our contributions can be summarized as follows:
\begin{itemize}
    \item In consideration of the distinctive geometric attributes of point cloud boundary points, this paper proposes a Boundary-Aware Graph attention network called BAGNet, which aims to concentrate the primary computational resources on boundary points with intricate features to improve the semantic segmentation speed and accuracy.
    \item BAGNet introduces a BAGLayer to capture contextual attention features from the boundary point by fusing the information of vertexes and edges in the graph and calculating attention coefficients.
    \item BAGNet adopts an attention pooling layer to aggregate the information of all the points and their neighbors to extract the global features.
    \item 
    We conduct extensive experiments on standard datasets to evaluate the performance of BAGNet. The results demonstrate that BAGNet outperforms state-of-the-art (SOTA) methods on model accuracy and inference time.
\end{itemize}

\section{Related Works}
\subsection{MLP-Based Methods}
In recent years, researchers have developed numerous methods to tackle the task of semantic segmentation on 3D point clouds. Some of them use MLP as the core building block, with a symmetric function for feature aggregation. Pointnet \cite{qi2017pointnet} is the first model to propose this method. PointNet++ \cite{qi2017pointnet++} builds upon its predecessor by improving the recognition of fine-grained geometric patterns and enhancing generalization to more complex scenes. To further enrich spatial understanding, Engelmann et al. \cite{engelmann2017exploring} proposed two mechanisms to extend PointNet, incorporating larger-scale spatial context through the aggregation of inputs from neighboring regions and the use of learned point descriptors. Ye et al. \cite{ye20183d} propose a novel unstructured end-to-end approach, capturing local
structures at various densities by taking multi-scale neighborhoods into
account. Hu et al. propose RandLa-Net \cite{hu2020randla}, which uses random point sampling instead of more
complex point selection approaches and designs a local feature aggregation module to increase the receptive field for each 3D point.

\subsection{Point Convolution-Based Methods}
Some researchers adapt convolution kernels to the underlying local geometries due to the disorder of point clouds. Atzmon et al. \cite{atzmon2018point} first propose a point convolutional neural network (PCNN), which maps the point cloud functions to the volumetric functions using the extension operator and maps it back using the restriction operator. Hua et al. \cite{hua2018pointwise} introduce a new operator called pointwise, which can be applied at each point of the point cloud. Komarichev et al. \cite{komarichev2019cnn} introduce annular convolution, a novel approach that enables the direct definition and computation of convolution operations on 3D point clouds. 
Li et al. \cite{li2018pointcnn} introduce an approach that learns an X-transformation from the input points, aiming to simultaneously assign weights to the corresponding features and reorder the points into a latent order.
A subset of point convolution-based methods employ graph convolutions \cite{kipf2017semi} to propagate and aggregate features across the graph. For instance, Lei et al. \cite{lei2020spherical} introduced a spherical kernel that enables graph neural networks to operate without relying on edge-dependent filter generation, offering improved computational efficiency for large-scale point clouds. Zhou et al. \cite{zhou2021adaptive} proposed an adaptive graph convolution approach, where kernels are dynamically generated based on the learned features of each point. Zhang et al. \cite{zhang2023improving}  adopt graph convolutions to aggregate local features in the shallow encoder stages.

\subsection{Attention Mechanism-Based Methods}
Attention mechanisms \cite{vaswani2017attention} are also incorporated to model long-range dependencies, suitable for handling point cloud irregularities. For example, Lai et al. \cite{lai2022stratified} proposed a stratified transformer capable of capturing long-range contextual dependencies, exhibiting strong generalization and high segmentation performance. Nie et al. \cite{nie2022pyramid} designed a scale pyramid architecture that integrates multi-scale features across different network stages for improved semantic segmentation. Park et al. \cite{park2022fast} introduced a fast point transformer, which employs a lightweight self-attention mechanism for encoding continuous 3D coordinates and leverages voxel hashing to enhance computational efficiency. Ran et al. \cite{ran2021learning} explore the possibilities of local relation operators and survey their feasibility. Zhang et al. \cite{zhang2022patchformer} proposed a patch-attention mechanism that adaptively learns a compact set of basis vectors to compute attention maps more efficiently. Zhao et al. \cite{zhao2021point} designed specialized self-attention layers tailored for point cloud data and constructed end-to-end self-attention networks based on them. To address weak supervision, Pan et al. \cite{pan2024less} introduced a novel label recommendation framework for point cloud semantic segmentation under limited annotation.

\begin{figure}
    \centering
    \includegraphics[width=0.48\textwidth]{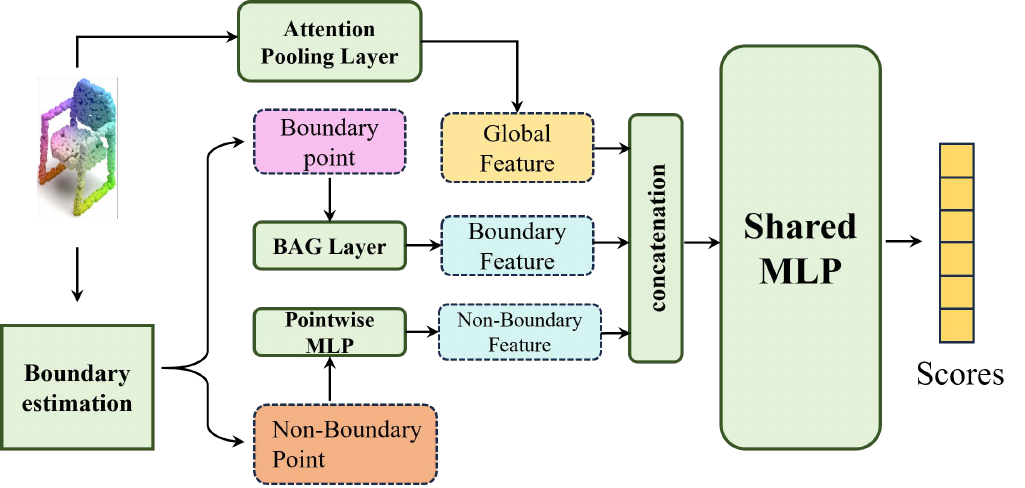}
    \caption{The overall architecture of our network: We divide the entire point cloud into boundary points and non-boundary points. We use the proposed BAGLayer to extract boundary features from boundary points, Pointwise MLP to extract simple non-boundary point features, and Attention Pooling Layer to extract global features from the raw point cloud.}
    \label{fig:overview}
\end{figure}

\section{Method}
This section begins with an overview of the network architecture, followed by a detailed description of the two core components of our approach: the BAGLayer and the attention pooling layer.

\begin{figure*}[t]
    \centering
    \includegraphics[width=1.0\linewidth]{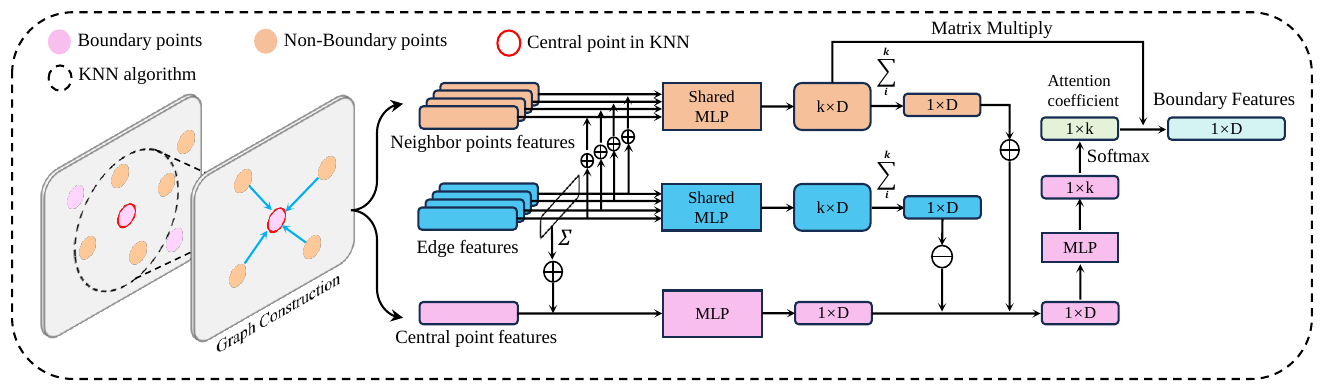}
    \caption{The architecture of the single channel BAGLayer: We select a boundary point as the central point and use the KNN algorithm to select $k$ neighbors around the central point to construct a graph. The features in the graph are divided into neighbor points features, edge features, and central point features.
    }
    \label{fig2}
\end{figure*}
\subsection{The Overall Architecture}
The overall architecture of our BAGNet is shown in Fig \ref{fig:overview}. Initially, we partition the complete point cloud into boundary points and non-boundary points according to boundary estimation based on normal vectors. As mentioned above, the
normal vector distribution of boundary points is
mostly chaotic, which is mainly reflected in two aspects: (1)  The normal vector of a boundary point shows a significant angular difference compared to the normal vectors of its neighboring points. (2)
The neighborhood of a boundary point tends to be unevenly distributed, often with fewer or no points on one side.
We employ the BAGLayer to manage boundary points and extract boundary features. We only use pointwise MLP to extract features for non-boundary points that are easy to identify. In order to further improve model accuracy, we use the attention pooling layer to extract the global feature of the raw point cloud. Then, these feature vectors are concatenated to obtain the final semantic segmentation result through a shared MLP.

\subsection{Boundary-Aware Graph attention Layer} 
Traditional methods either use several groups of shared MLPs to extract features, leading to a weak correlation between the boundary points and their neighbors, or construct a graph for each point in the point cloud, costing a lot of computation.
Our BAGLayer represents the connectivity between boundary points and their neighbors as a graph and fuses the information of vertexes and edges to enhance the relevance. BAGLayer learns attention coefficients for each boundary point, allowing it to assign varying levels of importance to different neighboring points. The complete architecture of the BAGLayer is illustrated in Figure \ref{fig2}.

\noindent\textbf{Graph Construction.} We model the point cloud as a graph to strengthen the relationships between boundary points and their neighboring points. Consider a graph $G_i = \left\{V_i,E_i \right\}$ constructed from a boundary point $\bar{p}_i$. We first find the $k$ nearest neighbors (KNN) of the boundary point $\bar{p}_i$, then $V_i$ can be represented as $V_i = \left\{\bar{p}_i, \bar{p}_{i1}, \bar{p}_{i2}, \ldots, \bar{p}_{ik}\right\}$, where $\bar{p}_{ij}$ is the $j$-th neighbor of $\bar{p}_i$. As for each edge in the graph, $E_i = \left\{(\bar{p}_i,\bar{p}_{ij}) | j = 1,2, \ldots, k \right\}$. To highlight the connection between two points, we set the weight of each edge as $\bar{W}_{ij} = \ln({ | \bar{p}_i - \bar{p}_{ij}}|)$, where the logarithm is taken to eliminate the linear relationship.

\noindent\textbf{Edge Vertex Fusion.} To fuse the information of boundary points and their neighbors as much as possible, our BAGLayer divides the graph structure into three parts: central point features (boundary point), edge features, and neighbor points features. In addition, we enhance the relevance of the boundary points through the method of edge vertex fusion. We add all the weights of the $k$ edges to their central point as follows:
\begin{equation}\label{eq1}
    \hat{p}_i = \bar{p}_i + \sum_{j=1}^{k} \bar{W}_{ij}
\end{equation}
For the $j$-th neighbor, we add the weight of its connected edges:
\begin{equation}\label{eq2}
    \hat{p}_{ij} = \bar{p}_{ij} + \bar{W}_{ij}  \quad j=1,2,\ldots,k
\end{equation}
After fusing the information of edges and vertexes, we encode the central point, neighbors, and edges by three different MLPs to get $\hat{p}_i'$, ,$\hat{p}_{ij}'$ and $\bar{W}_{ij}'$, respectively.
Since we have added the weight of the edges to the central point and the neighbors, the weight information is fused twice, so we need to reduce the impact of the edges as defined by the following equations:
\begin{equation}\label{eq10}
    y_i' = h(\hat{p}_i' + \sum_{j=1}^{k}\hat{p}_{ij}' - \sum_{j=1}^{k}\bar{W}_{ij}')
\end{equation}

\begin{table*}[t]
        \small 
	\caption{mIoU results of semantic segmentation on the ShapeNet part dataset.} 
	\label{tab 1} 
	\centering 
    \resizebox{\textwidth}{!}{

	\begin{tabular}{c c c c c c c c c c c c c c c c c c} 
		\toprule 
		  Model & avg & air. & bag & cap & car & cha. & ear. & gui. & kni. & lam. & lap. & mot. & mug & pis. & roc. & ska. & tab.\\ 
		\midrule 
            Kd-Net\cite{klokov2017escape} & 82.3 & 82.3 & 74.6 & 74.3 & 70.3 & 88.6 & 73.5 & 90.2 & 87.2 & 81.0 & 94.9 & 57.4 & 86.7 & 78.1 & 51.8 & 69.9 & 80.3\\
            PointNet\cite{qi2017pointnet} & 83.7 & 83.4 & 78.7 & 82.5 & 74.9 & 89.6 & 73.0 & 91.5 & 85.9 & 80.8 & 95.3 & 65.2 & 93.0 & 81.2 & 57.9 & 72.8 & 80.6\\
            PointNet++\cite{qi2017pointnet++} & 85.1 & 82.4 & 79.0 & 87.7 & 77.3 & 90.8 & 71.8 & 91.0 & 85.9 & 83.7 & 95.3 & 71.6 & 94.1 & 81.3 & 58.7 & 76.4 & 82.6\\
            RSNet\cite{huang2018recurrent} & 84.9 & 82.7 & 86.4 & 84.1 & 78.2 & 90.4 & 69.3 & 91.4 & 87.0 & 83.5 & 95.4 & 66.0 & 92.6 & 81.8 & 56.1 & 75.8 & 82.2\\
            SGPN\cite{wang2018sgpn} & 85.8 & 80.4 & 78.6 & 78.8 & 71.5 & 88.6 & 78.0 & 90.9 & 83.0 & 78.8 & 95.8 & \textbf{77.8} & 93.8 & 87.4 & 60.1 & \textbf{92.3} & 89.4\\
            DGCNN\cite{wang2019dynamic} & 85.1 & 84.2 & 83.7 & 84.4 & 77.1 & 90.9 & 78.5 & 91.5 & 87.3 & 82.9 & 96.0 & 67.8 & 93.3 & 82.6 & 59.7 & 75.5 & 82.0\\
            GAPointNet\cite{chen2021gapointnet} & 84.9 & 84.0 & 86.2 & 88.8 & \textbf{78.3} & 90.7 & 70.4 & 91.3 & 87.3 & 82.8 & 96.0 & 68.7 & \textbf{95.1} & 82.0 & 63.0 & 74.8 & 81.4\\

            PatchFormer \cite{zhang2022patchformer} & 84.3 & 84.0 & 86.4 & 85.2 & 77.6 & 90.8 & 72.3 & 90.7 & 87.1 & 82.4 & 96.2 & 75.1 & 93.6 & 82.3 & 60.2 & 83.4 & 88.2\\

            AF-GCN \cite{zhang2023improving} & 
            85.1 &
            84.2 &
            86.3 &
            87.4 &
            78.0 &
            90.5 &
            \textbf{79.2} &
            90.6 &
            82.9 &
            83.1 &
            96.0 &
            75.6 &
            94.1 &
            86.3 &
            61.2 &
            89.8 &
            87.4 \\
            Less is More\cite{pan2024less} &
            85.5 &
            83.9 &
            86.4 &
            80.1 &
            77.6 &
            91.2 &
            79.0 &
            91.4 &
            86.8 &
            83.2 &
            96.1 &
            76.7 &
            94.1 &
            86.8 &
            62.2 &
            86.0 &
            88.1 \\
            \midrule

            BAGNet & \textbf{86.2} & \textbf{84.5} & \textbf{86.9} & \textbf{89.0} & \textbf{78.3} & \textbf{92.2} & \textbf{79.2} & \textbf{91.8} & \textbf{87.4} & \textbf{83.9} & \textbf{97.7} & 75.3 &
            93.7 &
            \textbf{87.5} &
            \textbf{63.3} & 86.3 & \textbf{90.5}\\
		\bottomrule 
	\end{tabular}
    }
\end{table*}

\noindent\textbf{Attention Coefficient.} A set of attention coefficients is required to assign varying levels of importance to different neighboring points. A softmax function is also required to normalize the attention coefficient to make the neighborhood attention coefficients of different points consistent. We define our attention coefficients as follows:
\begin{equation}\label{eq12}
    a_{ij} = \frac{e^{y'_{ij}}}{\sum_{j}^{k}e^{y'_{ij}} }
\end{equation}
where $y'_{ij}$ is the $j_{th}$ component of $y'_i$. With the attention coefficients, we can obtain the boundary features that integrate their neighbors' information by matrix multiplication:
\begin{equation}\label{eq13}
    \hat{y}_i' = \sum_{j}^{k}a_{ij}\hat{p}'_{ij}
\end{equation}
where $k$ is the total number of points in the neighborhood and $\hat{y}_i'$ is the boundary features obtained through our BAGLayer for the boundary point $\bar{p}_i$.

\subsection{Attention Pooling Layer}
Although we mainly focus on modeling boundary points, we have not neglected the global features of the point cloud, as they are essential for maintaining model accuracy. To this end, we design an efficient method for extracting the global features of the point cloud, i.e., the attention pooling layer. The attention pooling layer considers the information of all points and their neighbors and uses an attention pooling operation to extract effective global features. This layer is lightweight and does not consume too many computational resources.
\begin{figure}[ht]
    \centering
    \includegraphics[width=1.0\linewidth]{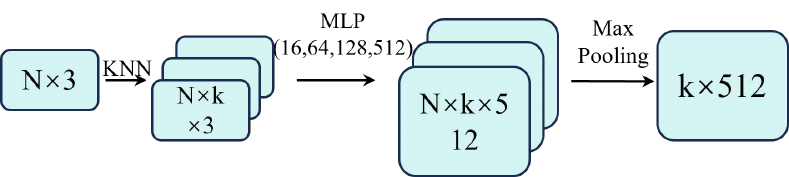}
    \caption{The architecture of the attention pooling layer.}
    \label{fig:pooling}
    \vspace{-5mm}
\end{figure}

Assume that there are \(N\) points in a point cloud, each point is defined by its 3D coordinates \((x,y,z)\). Then the input of the attention pooling layer has a shape of \((N,3)\). The first step in the pipeline involves passing the input data through a KNN network. The KNN network extracts the local geometric features of each point by considering its \(k\) nearest neighbors, thereby transforming the input data shape from \((N, 3)\) to \((N, k, 3)\), where \(k\) is the number of neighbors for each point. This step ensures that the local context of each point is preserved and encoded in the feature representation.

Next, the data is processed by an MLP with a fixed architecture of \((16, 64, 128, 512)\). The MLP progressively increases the dimensionality of the features, capturing more complex and higher-level representations of the point cloud. As a result, the data shape is further transformed from \((N, k, 3)\) to \((N, k, 512)\), where each point and its neighbors are now represented by a 512-dimensional feature vector. This high-dimensional feature space allows for richer and more discriminative feature extraction.

Finally, to aggregate the features across all points and generate a compact global representation, a max pooling operation is applied. The max pooling operation fuses the features of all points by selecting the maximum value along each feature dimension, resulting in a global feature vector with a shape of \((k, 512)\). This global feature encapsulates the most salient information from the entire point cloud, providing a comprehensive representation that is crucial for accurate semantic segmentation.

In summary, the attention pooling layer combines local feature extraction through KNN, feature enrichment via MLP, and global feature aggregation using max pooling to produce a robust and effective global representation of the point cloud. This global feature serves as a key component in the overall architecture, enabling the model to achieve high accuracy in point semantic segmentation tasks.

\section{Experiment}
In this section, we first compare the semantic segmentation performance of BAGNet with several SOTA point cloud segmentation methods on standard 3D point cloud datasets. Additionally, we conduct an ablation experiment to assess the effectiveness of the two important modules in BAGNet: BAGLayer and attention pooling layer.

\subsection{Experiment Setup}
\noindent\textbf{Dataset.} We evaluate the point cloud semantic segmentation performance on the widely used ShapeNet \cite{chang2015shapenet} and S3DIS \cite{armeni20163d} datasets. ShapeNet is a large-scale 3D model collection featuring more than 50,000 models across 55 object categories, including chairs, tables, cars, airplanes, and others. The S3DIS dataset, on the other hand, is a comprehensive resource for 3D semantic segmentation of indoor environments, containing over 272 million 3D points gathered from six expansive indoor areas, such as offices, conference rooms, and lounges.
\begin{table}
\centering
\caption{semantic segmentation results on S3DIS Area 5.}
\label{tab2}
\begin{tabular}{ccccc}
\toprule
\centering Model      & \centering mIoU & Overall Accuracy & Inference Time\\ 
\midrule
\centering SegCloud\cite{tchapmi2017segcloud}   & \centering 48.9\%   & \quad \quad  -     & 132.6ms           \\
\centering PointNet\cite{qi2017pointnet}   & \centering 47.6\%   & \quad \quad 78.5\%         & 92.4ms  \\
\centering DGCNN\cite{wang2019dynamic}      & \centering 56.1\%   & \quad \quad 84.1\%    & 362.1ms       \\
\centering GAPointNet\cite{chen2021gapointnet} & \centering 51.2\%   & \quad \quad \textbf{85.0}\%    & 163.8ms      \\
\centering AF-GCN\cite{zhang2023improving} & 
\centering 51.6\% & \quad \quad 83.7\% & 89.7ms \\
\midrule
\centering BAGNet       & \centering \textbf{56.5\%}        & \quad \quad 84.5\%          & \textbf{80.2ms}      \\ 
\bottomrule
\end{tabular}
\vspace{-5mm}
\end{table}

\noindent\textbf{Evaluation Metrics.} We utilize the mean Intersection over Union (mIoU) as the evaluation metric to assess the model's accuracy in performing the point cloud semantic segmentation task. Additionally, we measure inference time to evaluate the computation speed.

\noindent\textbf{Baseline Methods.} We use several SOTA point cloud segmentation methods as the baseline methods, including MLP-based methods (Kd-Net \cite{klokov2017escape}, PointNet \cite{qi2017pointnet}, PointNet++ \cite{qi2017pointnet++}, SegCloud \cite{tchapmi2017segcloud}), point convolution-based methods (RSNet \cite{huang2018recurrent}, SGPN \cite{wang2018sgpn}, DGCNN \cite{wang2019dynamic}, GAPointNet \cite{chen2021gapointnet}), and attention mechanism-based methods (PatchFormer \cite{zhang2022patchformer}, AF-GCN \cite{zhang2023improving}, Less is More \cite{pan2024less}). 

\noindent\textbf{Implementation Details}
In our attention pooling layer, we employ max pooling as the attention pooling function to enhance the network's overall robustness. For feature extraction of non-boundary points, we utilize PointNet++\cite{qi2017pointnet++} in the pointwise MLP, with the neighbor count parameter k set to 32. The BAGLayer's output dimension D was configured at 256. For training, we initialized the learning rate at 0.001 and applied a halving schedule every 40 epochs until reaching 0.00001, while maintaining a batch size of 16 throughout.

\subsection{Comparsion Experiment Results} 

\noindent\textbf{Accuracy.}
The comparison experiment we conduct on the dataset ShapeNet is shown in Table \ref{tab 1}. The results show that our framework outperforms SOTA methods on most of the objects of the ShapeNet dataset. On some objects with many boundary edges, such as laptops, tables, and chairs, our model has shown a greater improvement compared with other models. On these three types of objects, the accuracy of BAGNet is respectively 1\%, 1.6\%, and 1.1\% higher than the SOTA methods. This is because we enhance the connection between two points in the graph through edge-vertex fusion. Besides, our method achieves the highest average accuracy compared to other SOTA methods. These results indicate that our method not only performs well on objects with clear boundaries but also generally exhibits good performance on regular objects, making it a highly competitive model in this field. 

\begin{figure}[ht]
    \centering
    \includegraphics[width=0.48\textwidth]{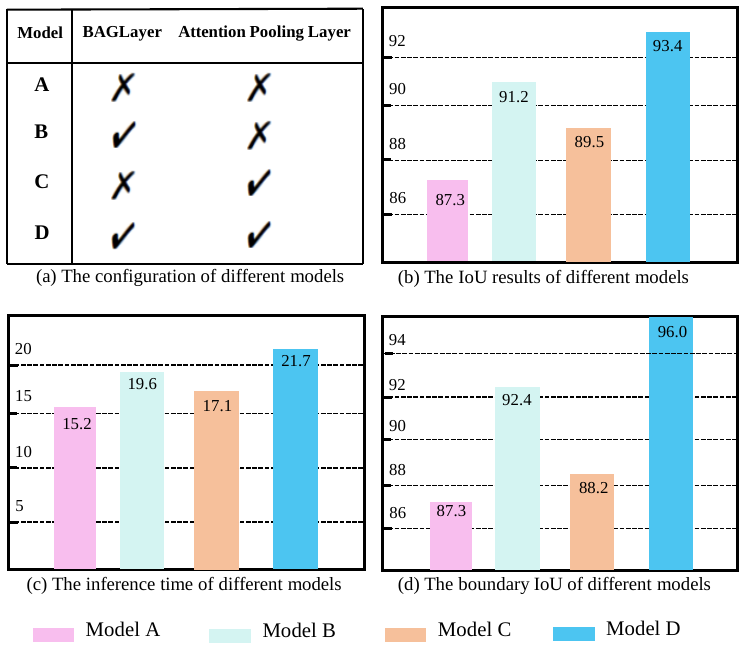}
    \caption{The semantic segmentation results of different models on the dataset ShapeNet.}
    \vspace{-5mm}
    \label{fig:ablation}
    
\end{figure}
We also present the experimental results of our model on the S3DIS dataset area 5 in Table \ref{tab2}. As is shown in Table \ref{tab2}, our model has the best mIoU and performs better overall accuracy than all the other models except GAPointNet. In fact, GAPointNet also uses an attention mechanism to extract features, but in its semantic segmentation network architecture, it uses two GAPLayers with four heads and two attention pools to extract global features, which makes it much easier to identify these relatively larger scale parts. 

\noindent\textbf{Inference Time.} We compare the inference time of BAGNet with baseline methods on the S3DIS dataset area 5 in Table \ref{tab2}. From the table, we can see that BAGNet achieves the least inference time. Compared to other methods, BAGNet can reduce inference time by 13.2\% to 77.9\%. This is because we concentrate most of the computational resources on the boundary points while using lightweight processing for other non-boundary points and global features to maintain the necessary model accuracy.

\subsection{Ablation Study} 

To demonstrate the effectiveness of the two modules in our framework, we also test our model with different structure settings on the ShapeNet part dataset.

The different structural settings of the model are shown in Figure \ref{fig:ablation} (a). We conduct experiments on these four different models. The mIoU results are shown in Figure \ref{fig:ablation}(b). Our BAGLayer and the attention pooling layer improve the mIoU by about 3.9\% and 2.2\%, respectively. The combination of these two modules has improved the mIoU by 6.1\%. Figure \ref{fig:ablation} (c) illustrates the impact of the two modules on the inference time. 
The inference time of our model is only increased by 4.4ms and 1.9ms with these two modules, respectively. However, as we can see from Figure \ref{fig:ablation} (b), these two modules significantly improve the model accuracy, making the slight increase in inference time caused by introducing these modules acceptable. We also test the mIoU of the boundary points as shown in Figure \ref{fig:ablation}(d). The result shows that our BAGLayer and the attention pooling layer can improve the accuracy of boundary point semantic segmentation to varying degrees.


\section{Conclusion}
This paper introduces BAGNet, a boundary-aware graph attention network designed for semantic segmentation of 3D point clouds. BAGNet designs BAGLayer to concentrate most of the computation resources on boundary points. BAGLayer utilizes the fusion of edge-vertex information within the graph to capture contextual attention features and enables each boundary point to calculate a unique set of attention coefficients, which encapsulates the connectivity and relational data between the boundary point and all its adjacent neighbors. BAGNet further integrates a computationally efficient attention pooling layer to capture global point cloud representations without compromising segmentation accuracy. Experimental results obtained show the effectiveness of BAGNet in improving boundary point accuracy and reducing computational load. 

\section{Acknowledgements}
This work was supported by the National Key Research and Development Program of China (Youth Scientist Project) under Grant No. 2024YFB4504300 and the Shenzhen-Hong Kong Joint Funding Project (Category A) under Grant No. SGDX20240115103359001.

\bibliographystyle{ieeetr}
\bibliography{strings}

\end{document}